\documentclass{article}
\usepackage{ijcai25}

\usepackage{times}
\usepackage{soul}
\usepackage{xcolor}
\usepackage{url}
\usepackage[most]{tcolorbox}
\usepackage{tcolorbox}
\usepackage[hidelinks]{hyperref}
\usepackage[utf8]{inputenc}
\usepackage[small]{caption}
\usepackage{graphicx}
\usepackage{amsmath}
\usepackage{amsthm}
\usepackage{booktabs}
\usepackage{algorithm}
\usepackage{algorithmic}
\usepackage[switch]{lineno}

\usepackage{prettyref}
\newcommand{\pref}{\prettyref}
\definecolor{note}{RGB}{238, 248, 248}

\title{Conversational Exploration of Literature Landscape with LitChat}

\author{
Mingyu Huang$^1$
\and
Shasha Zhou$^2$\and
Yuxuan Chen$^{3}$\And
Ke Li$^1$\\
\affiliations
$^1$College of Computer Science and Engineering, University of Electronic Science and Technology of China, Xiyuan Avenue 2006, Chengdu, China\\
$^2$Department of Computer Science, University of Exeter, EX4 4QF, Exeter, United Kingdom\\
$^3$James Watt School of Engineering, University of Glasgow, G12 8QQ, Glasgow, United Kindom\\
\emails
m.huang.gla@outlook.com,
\{sz484, k.li\}@exeter.ac.uk,
2959902c@student.gla.ac.uk
}

\begin{document}

\maketitle

\begin{abstract}
    We are living in an era of ``big literature'', where the volume of digital scientific publications is growing exponentially. While offering new opportunities, this also poses challenges for understanding literature landscapes, as traditional manual reviewing is no longer feasible. Recent large language models (LLMs) have shown strong capabilities for literature comprehension, yet they are incapable of offering “comprehensive, objective, open and transparent” views desired by systematic reviews due to their limited context windows and trust issues like hallucinations. Here we present LitChat, an end-to-end, interactive and conversational literature agent that augments LLM agents with data-driven discovery tools to facilitate literature exploration. LitChat automatically interprets user queries, retrieves relevant sources, constructs knowledge graphs, and employs diverse data-mining techniques to generate evidence-based insights addressing user needs. We illustrate the effectiveness of LitChat via a case study on AI4Health, highlighting its capacity to quickly navigate the users through large-scale literature landscape with data-based evidence that is otherwise infeasible with traditional means.
\end{abstract}

\section{Introduction}
\label{sec:introduction}

The past decade has witnessed an unprecedented expansion of scientific literature on many topics. This deluge of digital data on publications offers unprecedented opportunities for scholars and practitioners to explore patterns characterizing the structure and evolution of the underlying literature landscapes while simultaneously confronting them with new challenges. Scholars often develop systematic literature reviews (SLRs)~\cite{KitchenhamBBTBL09} to obtain comprehensive overviews of the relevant topics by manually reviewing a set of relevant publications. However, the exponentially increasing volume of literature means that the provision of ``comprehensive, objective, open and transparent'' (COOT) principles is no longer feasible by traditional means. For example, the number of publications on the applications of artificial intelligence in healthcare has reached 100,000 in 2023, which is far beyond the capacity of human experts to review.

Large language models (LLMs)~\cite{GPT4} have now emerged as transformative tools that promise to revolutionize scientific literature exploration and understanding. Autosurvey~\cite{Autosurvey} is a recent example of such a system that uses LLMs to automatically generate literature reviews. Another system, PaperQA2~\cite{Skarlinski24}, leverages LLMs and retrieval-augmented generation (RAG)~\cite{RAG} to answer questions about scientific publications. While they have shown some promise, they also fail to provide COOT views of literature landscapes. For example, due to the limited context windows of LLMs, it is infeasible to feed the entire literature corpus, which may contain thousands of papers, into the LLMs. This is in contrast to the ``comprehensive'' principle of SLR. Similarly, as probablistic models, LLMs function as black boxes and are prone to hallucinations~\cite{JiLFYSXIBMF23}, which may lead to incorrect or misleading information that violates the ``objective'' and ``transparent'' principles. 

Long before the emergence of LLMs, scholars have explored the use of data-driven methods to generate insights that can guide literature exploration. For example, ~\citeauthor{AI4Life} leverages topic modeling to explore latent themes in large corpora of literature on the use of artificial intelligence in life science research. ~\citeauthor{JeongZD23} uses community detection techniques to explore key research topics in authentication. Our recent work~\cite{HuangL24} combines these with author collaboration network analysis to provide a holistic view on the research landscape of multi-objective optimization research. These methods could provide quantitative evidence to facilitate SLR under the COOT principles.

Inspired by this, here we present LitChat, an end-to-end, interactive and conversational literature agent that combines LLM agents and data-driven discovery tools to facilitate literature exploration. LitChat can automatically interpret user queries, design search queries, retrieves relevant sources from online databases, constructs bibliographic knowledge graphs, and employs diverse data-mining techniques to deliver evidence-based insights addressing user needs. It is able to perform explorations of different granularities, from providing an overview of the literature landscape to drilling down into the details of individual papers or researchers. We demonstrate the effectiveness of LitChat through a case study on AI4Health, highlighting its capacity to guide users through a timely, large-scale and rapidly evolving research landscape.

\section{LitChat}
\label{sec:litchat}

\begin{figure}[t!]
    \centering
    \includegraphics[width=\linewidth]{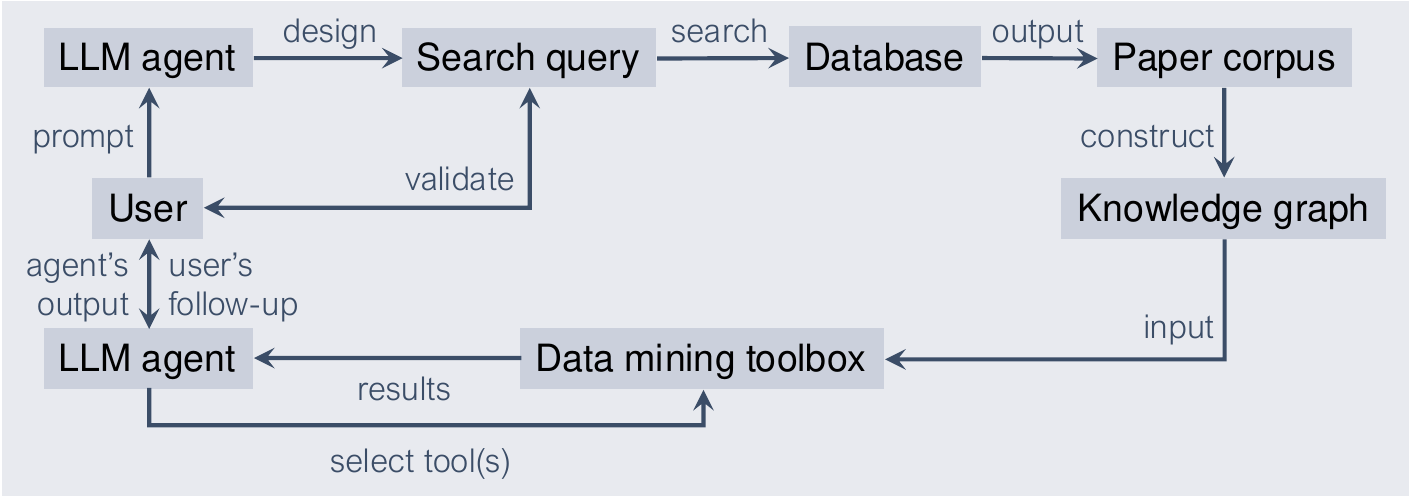}
    \caption{Schematic overview of LitChat.}
    \label{fig:system}
\end{figure}

\begin{figure*}[t!]
    \centering
    \includegraphics[width=\linewidth]{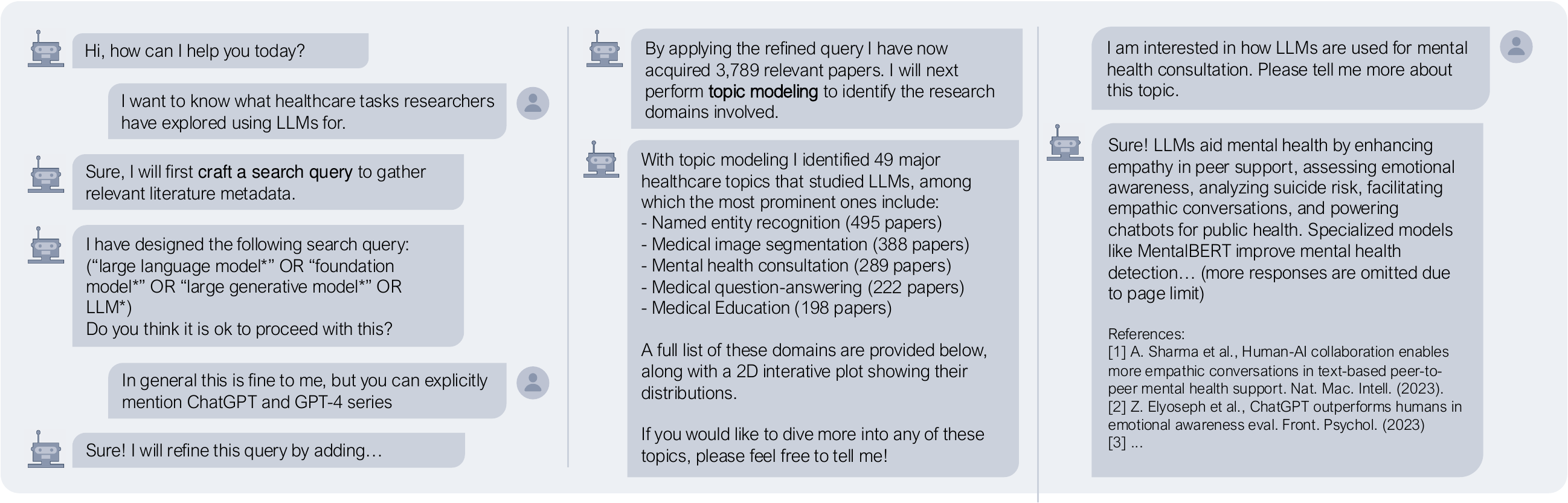}
    \caption{An example of LitChat in action.}
    \label{fig:demo}
\end{figure*}

A schematic overview of LitChat is depicted in~\pref{fig:system}. Overall, LitChat consists of two LLM agents, both of which can interact with users in a conversational manner. The first agent (upper) takes user queries and devises boolean search queries to retrieve relevant publications from online databases. The collected papers along with metadata are then used to construct a bibliographic knowledge graph (BKG). The second agent (lower) then selects appropriate data-mining techniques to explore the obtained BKG and generate responses to user queries based on the analysis results. 

At a higher level, the whole system is connected to a frontend built using JavaScript and React, which allows users to interact through a web interface. This frontend is connected to a backend using a Python Flask web app. The backend holds the LLM agents, the BKG (via Neo4j graph database), the paper metadata (via SQLite databases), and the data mining library. Below we delineate each component of LitChat.

\subsection{Query Design and Data Acquisition} 

\paragraph{Initial query design.} When a user enters a query, e.g., ``I want to know what healthcare tasks researchers have explored using LLMs for'' (see~\pref{fig:demo}), the first LLM agent identifies the key research domains involved and synthesizes them into a boolean search query that can be used on established databases like Web of Science\footnote{https://www.webofscience.com}, Scopus\footnote{https://www.scopus.com}, or Semantic Scholar\footnote{https://www.semanticscholar.org/}. Benefiting from the extensive training on large-scale corpora, LLMs have shown human-level performance in query formulation~\cite{WangSKZ23}. To further ensure the quality of the designed query, 10 in-context-learning examples are provided to the LLM agent. These examples are extracted from established SLRs from diverse domains in which domain experts explicitly describe the search queries they used for literature search.

\begin{tcolorbox}[breakable, title after break=, height fixed for = none, colback = note, boxrule = 0pt, sharpish corners, top = 4pt, bottom = 4pt, left = 4pt, right = 4pt, toptitle=2pt, bottomtitle=2pt]
    \small \textbf{Note on the choice of LLMs:}  Unlike other LLM-based literature agents like Autosurvey and GraphRAG, which prompt LLMs to scrutinize the entire text corpus before reaching the final outputs, LitChat only uses LLMs to directly interact with the users for essential communications like discussing the query, explaining the results, and answering questions. This makes LitChat significantly more ``token-efficient'' and thus more affordable for broader applications. We thereby choose GPT-4o to enhance performance, though any other LLMs can also be used.
\end{tcolorbox}

\paragraph{Interactive refinement.} The initial query will then be sent to the user for confirmation. The user can either approve it or provide feedback to refine, e.g., by adding or removing keywords. In the latter case, the LLM agent will update the query accordingly and ask for confirmation again. This process continues until the user is satisfied with the query.

\paragraph{Database search and metadata.} Once the query is confirmed, the LLM agent forwards it to an online database via API calls to retrieve relevant publications. Here we use the WoS API for this purpose, which is one of the most comprehensive literature databases and a common choice for SLR. The WoS API returns the matched publications in JSON format, which consists of rich curated metadata such as abstract, authors, venues, publication date, citations. The agent stores this information in a SQLite database for further processing. 

\begin{tcolorbox}[breakable, title after break=, height fixed for = none, colback = note, boxrule = 0pt, sharpish corners, top = 4pt, bottom = 4pt, left = 4pt, right = 4pt, toptitle=2pt, bottomtitle=2pt]
    \small \textbf{Note on full paper data:} Due to copyright issues, LitChat typically only retrieves paper metadata from online databases, unless the full paper is open access. Yet, a user is able to upload PDF documents to LitChat for further analysis. 
\end{tcolorbox}

\subsection{Bibliographic Knowledge Graph Construction}

Once the paper metadata is collected, LitChat parses it to construct a BKG, which essentially contains all the information regarding the underlying literature landscape. Notably, this graph is heterogeneous, as it consists of different types of entities, e.g., papers, authors, venues, keywords, institutions, etc., and complex relationships among them (e.g., citation, co-authorship, keyword co-occurrence, etc.). Paper abstracts are also integrated as part of node attributes, and we generate embeddings for them using the \texttt{voyage-3-large}\footnote{https://blog.voyageai.com/2025/01/07/voyage-3-large/} model, a top performer on the MTEB leaderboard. The final BKG provides the basis for all the subsequent data mining and response generation tasks. Upon construction, the BKG will be stored in a Neo4j graph database to allow for efficient querying and analysis.

\subsection{Data Mining and Response Generation}

After obtaining the BKG, the second LLM agent, also based on GPT-4o, would take another look at the user query and select one or more data mining approaches from a predefined library that can be helpful for answering the query. While an exhaustive list of data mining techniques is not feasible here, some important ones include:

\begin{itemize}
    \item \textbf{Topics and keywords:} Applying topic modeling~\cite{Maarten} to uncover latent research themes and their temporal evolution in the BKG; Traverse keyword co-occurrence networks~\cite{Miao22} to identify interdisciplinary research topics. 
    \item \textbf{Papers and citations:} Exploiting the topology of paper citation network to identify or predict impactful works~\cite{Weis} and recommend similar papers~\cite{BoyackK10}.
    \item \textbf{Authors and research groups:} Exploring author collaboration networks to identify active researchers and groups with network metrics~\cite{Palla07}. 
    \item \textbf{Scientific discovery:} Learning hidden patterns embedded in the BKG to infer meaningful research ideas that transcend existing individual knowledge and cross-domain boundaries~\cite{Krenn23,Bai24}.
\end{itemize}

After LitChat executes a tool, it composes the results of the operations into a natural language response that it returns to the user. LitChat generates these responses by filling in templates associated with each operation based on the results. The responses also include sufficient context to understand the results and opportunities for following up. In addition to these advanced data-driven tools, LitChat also supports basic tasks like question-answering and summarization based on RAG techniques just as many other literature chatbots.

\section{Case Study on AI4Health}
\label{sec:case}

\begin{figure}[t!]
    \centering
    \includegraphics[width=\linewidth]{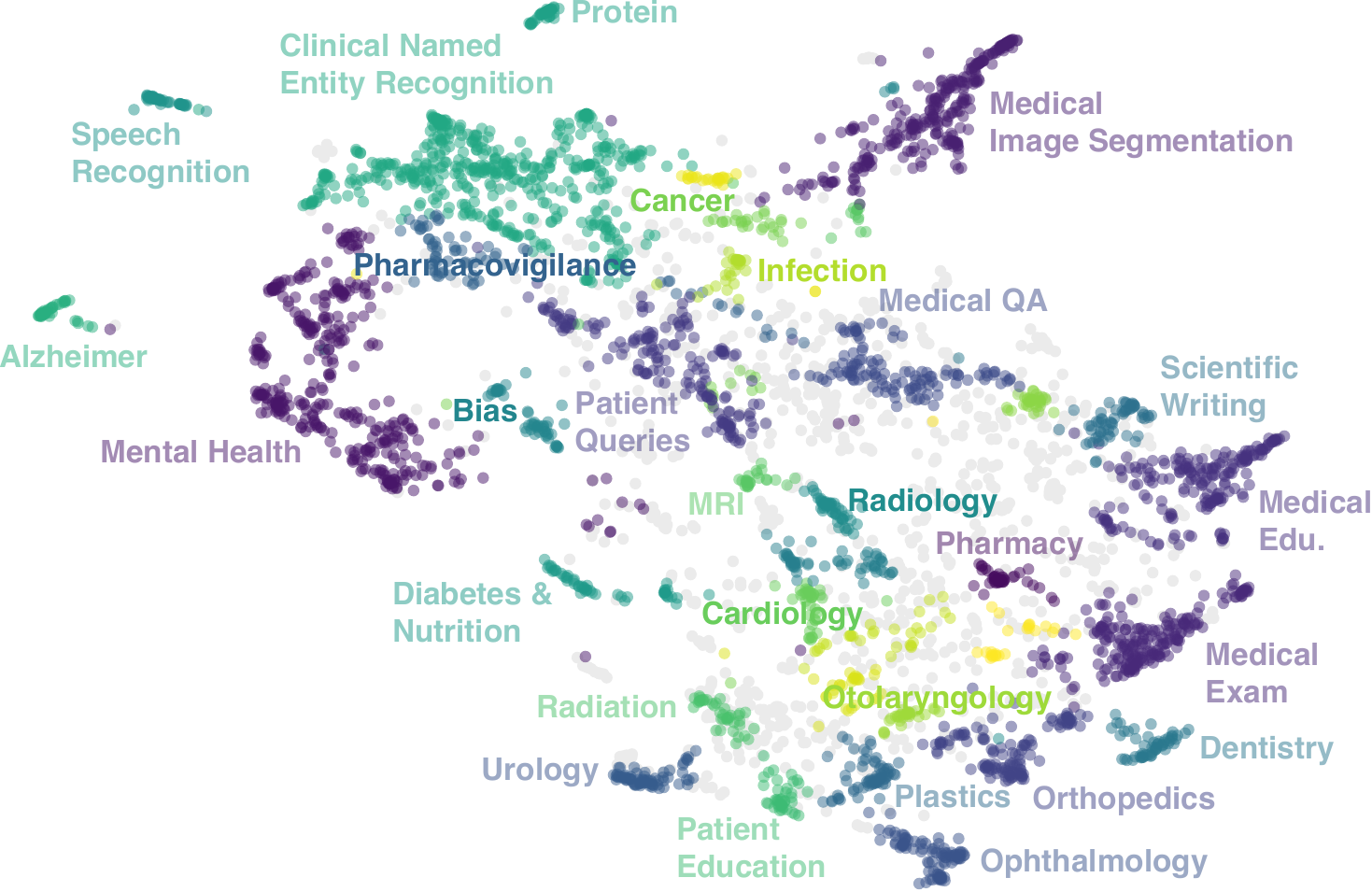}
    \caption{2D visualization of the research landscape of LLM applications in healthcare provided by LitChat. Each point represents a paper, colored by the research topic as identified by topic modeling. Light gray points indicate outliers that are not assigned to any topic.}
    \label{fig:topic2d}
\end{figure}

To demonstrate the effectiveness of LitChat, here we present a case study on the AI4Health domain. Specifically, as shown in~\pref{fig:demo}, the user first asks ``I want to know what healthcare tasks researchers have explored using LLMs for?'' As one can imagine, answering this requires a comprehensive audit of the relevant literature, which is beyond the capacity of human experts. To address this, LitChat first creates a candidate search query and engages with the user to refine it to allow for a comprehensive coverage of relevant literature. The final query is then sent to the Web of Science database, from which 3,700+ papers are retrieved and synthesized into a BKG. Given the large volume of papers, traditional LLM-based systems would struggle to process them all, and this is where LitChat's data-driven tools come into play.

Since the user is interested in an overview of certain research topics, LitChat chooses \texttt{topic\_modeling} to analyze the BKG. This function clusters the retrieved papers into different topics based on their semantic proximity (e.g., in the upper left region are topics related to brain and recognition, e.g., ``Alzheimer'', ``Mental Health'', and ``Speech Recognition''). LitChat then generates response by highlighting the most prominent topics, and also provides a 2D visualization of the literature landscape (see~\pref{fig:topic2d}). Note that in the response, LitChat is able to explicitly indicate the size of each topic as determined by the topic model, which conforms to the ``objective'' and ``transparent'' principles of SLR. 

When the user asks for more details on a specific topic (here ``mental health''), LitChat can use the \texttt{topic} tool to recommend representative papers, and generate topic summaries based on them. This conversation can continue to drill down into the details of one specific paper in this topic, historical trends of the topic, or the authors involved, etc., and each response would also be grounded on data-based evidence. 

\section{Conclusion}
\label{sec:conclusion}

In this paper we introduced LitChat, an interactive LLM agent that is augmented with data-driven discovery tools to aid the exploration of large-scale literature landscapes. With a case study on AI4Health, we demonstrated how LitChat can help users to quickly navigate the terrain of LLM applications in healthcare research in an end-to-end fashion, which is otherwise infeasible with traditional manual reviewing or pure LLM-based systems. We hope that LitChat opens up new possibilities for scholars and practitioners to understand scientific literature in this era of literature explosion.

\clearpage 
\bibliographystyle{named}
\bibliography{ijcai25}

\begin{thebibliography}{}

\bibitem[\protect\citeauthoryear{Bai \bgroup \em et al.\egroup }{2024}]{Bai24}
Jiaru Bai, Sebastian Mosbach, Connor~J. Taylor, Dogancan Karan, Kok~Foong Lee,
  Simon~D. Rihm, Jethro Akroyd, Alexei~A. Lapkin, and Markus Kraft.
\newblock A dynamic knowledge graph approach to distributed self-driving
  laboratories.
\newblock {\em Nat. Commun.}, 15(1):462, 2024.

\bibitem[\protect\citeauthoryear{Boyack and Klavans}{2010}]{BoyackK10}
Kevin~W. Boyack and Richard Klavans.
\newblock Co-citation analysis, bibliographic coupling, and direct citation:
  Which citation approach represents the research front most accurately?
\newblock {\em J. Assoc. Inf. Sci. Technol.}, 61(12):2389--2404, 2010.

\bibitem[\protect\citeauthoryear{Gao \bgroup \em et al.\egroup }{2023}]{RAG}
Yunfan Gao, Yun Xiong, Xinyu Gao, Kangxiang Jia, Jinliu Pan, Yuxi Bi, Yi~Dai,
  Jiawei Sun, Qianyu Guo, Meng Wang, and Haofen Wang.
\newblock Retrieval-augmented generation for large language models: {A} survey.
\newblock {\em CoRR}, abs/2312.10997, 2023.

\bibitem[\protect\citeauthoryear{Grootendorst}{2022}]{Maarten}
Maarten Grootendorst.
\newblock Bertopic: Neural topic modeling with a class-based {TF-IDF}
  procedure.
\newblock {\em CoRR}, abs/2203.05794, 2022.

\bibitem[\protect\citeauthoryear{Huang and Li}{2024}]{HuangL24}
Mingyu Huang and Ke~Li.
\newblock A survey of decomposition-based evolutionary multi-objective
  optimization: Part {II} - {A} data science perspective.
\newblock {\em CoRR}, abs/2404.14228, 2024.

\bibitem[\protect\citeauthoryear{Jeong \bgroup \em et al.\egroup
  }{2023}]{JeongZD23}
Jongkil~Jay Jeong, Yevhen Zolotavkin, and Robin Doss.
\newblock Examining the current status and emerging trends in continuous
  authentication technologies through citation network analysis.
\newblock {\em {ACM} Comput. Surv.}, 55(6):122:1--122:31, 2023.

\bibitem[\protect\citeauthoryear{Ji \bgroup \em et al.\egroup
  }{2023}]{JiLFYSXIBMF23}
Ziwei Ji, Nayeon Lee, Rita Frieske, Tiezheng Yu, Dan Su, Yan Xu, Etsuko Ishii,
  Yejin Bang, Andrea Madotto, and Pascale Fung.
\newblock Survey of hallucination in natural language generation.
\newblock {\em {ACM} Comput. Surv.}, 55(12):248:1--248:38, 2023.

\bibitem[\protect\citeauthoryear{Kitchenham \bgroup \em et al.\egroup
  }{2009}]{KitchenhamBBTBL09}
Barbara~A. Kitchenham, Pearl Brereton, David Budgen, Mark Turner, John Bailey,
  and Stephen~G. Linkman.
\newblock Systematic literature reviews in software engineering - {A}
  systematic literature review.
\newblock {\em Inf. Softw. Technol.}, 51(1):7--15, 2009.

\bibitem[\protect\citeauthoryear{Krenn \bgroup \em et al.\egroup
  }{2023}]{Krenn23}
Mario Krenn, Lorenzo Buffoni, Bruno Coutinho, Sagi Eppel, Jacob~Gates Foster,
  Andrew Gritsevskiy, Harlin Lee, Yichao Lu, Jo{\~a}o~P. Moutinho, Nima
  Sanjabi, Rishi Sonthalia, Ngoc~Mai Tran, Francisco Valente, Yangxinyu Xie,
  Rose Yu, and Michael Kopp.
\newblock Forecasting the future of artificial intelligence with machine
  learning-based link prediction in an exponentially growing knowledge network.
\newblock {\em Nat. Mac. Intell.}, 5(11):1326--1335, 2023.

\bibitem[\protect\citeauthoryear{Miao \bgroup \em et al.\egroup
  }{2022}]{Miao22}
Lili Miao, Dakota Murray, Woo-Sung Jung, Vincent Larivi{\`e}re, Cassidy~R.
  Sugimoto, and Yong-Yeol Ahn.
\newblock The latent structure of global scientific development.
\newblock {\em Nat. Hum. Behav.}, 6(9):1206--1217, 2022.

\bibitem[\protect\citeauthoryear{OpenAI}{2023}]{GPT4}
OpenAI.
\newblock {GPT-4} technical report.
\newblock {\em CoRR}, abs/2303.08774, 2023.

\bibitem[\protect\citeauthoryear{Palla \bgroup \em et al.\egroup
  }{2007}]{Palla07}
Gergely Palla, Albert-L{\'a}szl{\'o} Barab{\'a}si, and Tam{\'a}s Vicsek.
\newblock Quantifying social group evolution.
\newblock {\em Nature}, 446(7136):664--667, 2007.

\bibitem[\protect\citeauthoryear{Schmallenbach \bgroup \em et al.\egroup
  }{2024}]{AI4Life}
Leo Schmallenbach, Till~W. B{\"a}rnighausen, and Marc~J. Lerchenmueller.
\newblock The global geography of artificial intelligence in life science
  research.
\newblock {\em Nat. Commun.}, 15(1):7527, 2024.

\bibitem[\protect\citeauthoryear{Skarlinski \bgroup \em et al.\egroup
  }{2024}]{Skarlinski24}
Michael~D. Skarlinski, Sam Cox, Jon~M. Laurent, James~D. Braza, Michaela~M.
  Hinks, Michael~J. Hammerling, Manvitha Ponnapati, Samuel~G. Rodriques, and
  Andrew~D. White.
\newblock Language agents achieve superhuman synthesis of scientific knowledge.
\newblock {\em CoRR}, abs/2409.13740, 2024.

\bibitem[\protect\citeauthoryear{Wang \bgroup \em et al.\egroup
  }{2023}]{WangSKZ23}
Shuai Wang, Harrisen Scells, Bevan Koopman, and Guido Zuccon.
\newblock Can chatgpt write a good boolean query for systematic review
  literature search?
\newblock In {\em {SIGIR}'23: Proc. of the 46th International {ACM} {SIGIR}
  Conference on Research and Development in Information Retrieval}, pages
  1426--1436. {ACM}, 2023.

\bibitem[\protect\citeauthoryear{Wang \bgroup \em et al.\egroup
  }{2024}]{Autosurvey}
Yidong Wang, Qi~Guo, Wenjin Yao, Hongbo Zhang, Xin Zhang, Zhen Wu, Meishan
  Zhang, Xinyu Dai, Min zhang, Qingsong Wen, Wei Ye, Shikun Zhang, and Yue
  Zhang.
\newblock Autosurvey: Large language models can automatically write surveys.
\newblock In {\em {NeurIPS}'25: Proc. of the 38th Annual Conference on Neural
  Information Processing Systems}, 2024.

\bibitem[\protect\citeauthoryear{Weis and Jacobson}{2021}]{Weis}
James~W. Weis and Joseph~M. Jacobson.
\newblock Learning on knowledge graph dynamics provides an early warning of
  impactful research.
\newblock {\em Nat. Biotechnol.}, 39(10):1300--1307, 2021.

\end{thebibliography}

\end{document}